\documentclass{article}

\usepackage[preprint,nonatbib]{neurips_2022}

\usepackage[utf8]{inputenc} 
\usepackage[T1]{fontenc}    
\usepackage[pagebackref,breaklinks,colorlinks]{hyperref}      
\usepackage{url}            
\usepackage{booktabs}       
\usepackage{amsfonts}       
\usepackage{nicefrac}       
\usepackage{microtype}      
\usepackage{xcolor}         
\usepackage{graphicx}
\usepackage{authblk}

\title{Characterizing information loss in a chaotic double pendulum with the Information Bottleneck}

\author[1]{Kieran A. Murphy}
\author[1,2,3,4,5,6,7]{Dani S. Bassett}

\affil[1]{Dept. of Bioengineering, School of Engineering \& Applied Science, \linebreak U. of Pennsylvania, Philadelphia, PA 19104, USA}
\affil[2]{Dept. of Electrical \& Systems Engineering, School of Engineering \& Applied Science,\linebreak U. of Pennsylvania, Philadelphia, PA 19104, USA}
\affil[3]{Dept. of Neurology, Perelman School of Medicine, U. of Pennsylvania, Philadelphia, PA 19104, USA}
\affil[4]{Dept. of Psychiatry, Perelman School of Medicine, U. of Pennsylvania, Philadelphia, PA 19104, USA}
\affil[5]{Dept. of Physics \& Astronomy, College of Arts \& Sciences, U. of Pennsylvania, Philadelphia, PA 19104, USA}
\affil[6]{The Santa Fe Institute, Santa Fe, NM 87501, USA}
\affil[7]{To whom correspondence should be addressed: dsb@seas.upenn.edu}

\begin{document}

\maketitle

\vspace{-7mm}

\begin{abstract}
A hallmark of chaotic dynamics is the loss of information with time. Although information loss is often expressed through a connection to Lyapunov exponents---valid in the limit of high information about the system state---this picture misses the rich spectrum of information decay across different levels of granularity. Here we show how machine learning presents new opportunities for the study of information loss in chaotic dynamics, with a double pendulum serving as a model system. We use the Information Bottleneck as a training objective for a neural network to extract information from the state of the system that is optimally predictive of the future state after a prescribed time horizon. We then decompose the optimally predictive information by distributing a bottleneck to each state variable, recovering the relative importance of the variables in determining future evolution.
The framework we develop is broadly applicable to chaotic systems and pragmatic to apply, leveraging data and machine learning to monitor the limits of predictability and map out the loss of information. 
\end{abstract}

\vspace{-4.5mm}
\section{Introduction}
\vspace{-2mm}
A fundamental aspect of chaos is the loss of information over time: for any measurement of a chaotic system with finite resolution, there is a finite time horizon beyond which the measurement bears no predictive power~\cite{shaw1981,farmer1982informationdimension,wolf1985Lyapunov,boffetta2002predictability,james2014chaosforgets,berera2019turbulence}.
The premise of this work is simple: to find the optimally predictive information in a chaotic system at different levels of granularity, and to study how the predictive power of this information erodes with the passage of time.

Information loss is intimately connected to the distortion of regions in phase space by chaotic dynamics, and thus to Lyapunov exponents: the sum of the positive Lyapunov exponents gives the Kolmogorov-Sinai (KS) entropy, the average rate of information loss\cite{boffetta2002predictability}.
However, these quantities are valid in the limit of maximal information---where infinitesimally-separated trajectories are discernible---and are thus somewhat removed from reality~\cite{boffetta2002predictability}.
The $(\epsilon, \tau)$-entropy generalizes the KS entropy, describing the loss of predictive power for different amounts of information about the system state~\cite{gaspard1993noise}.
It is defined by way of a rate-distortion objective that minimizes the rate of information needed to predict the system state better than a threshold value of some chosen measure of distortion.

The Information Bottleneck (IB) is a rate-distortion problem where the measure of distortion is based on mutual information; it extracts the information from one variable that is most shared with a second variable~\cite{tishbyIB2000}.
We can use the IB to find optimally predictive information from one state of a chaotic system about a future state, and at the same time measure the loss of predictive power~\cite{creutzig2009pastfutureIB}.
In this work we develop a framework that uses the IB for analyzing chaotic dynamics with machine learning.  
We use an interpretable variant of the IB, the Distributed IB~\cite{dib}, to decompose the optimally predictive information in terms of a system's state variables.

\begin{figure}
    \centering
    \includegraphics[width=\linewidth]{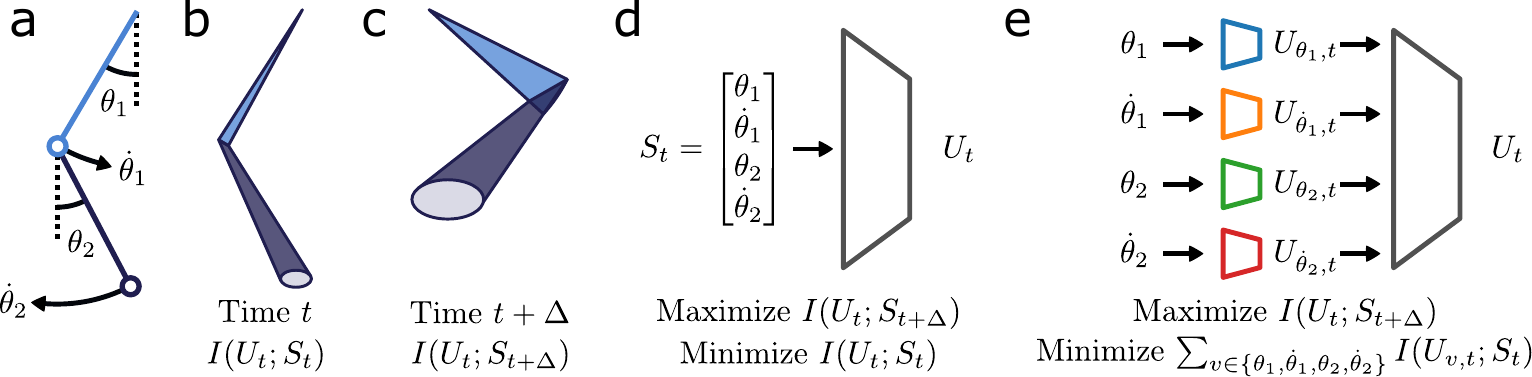}
    \caption{\textbf{Predicting the future of a double pendulum given finite information.}
    \textbf{(a)} The double pendulum system with four state variables $\theta_1, \dot{\theta}_1, \theta_2, \dot{\theta}_2$.
    \textbf{(b)} At time $t$ the state $S_t$ is measured, yielding a random variable $U_t$. The finite mutual information between the state and the measurement $I(S_t;U_t)$ manifests as a continuum of underlying states that cannot be discerned given an outcome of $U_t$.
    \textbf{(c)} Information is lost over time due to the double pendulum's chaotic dynamics. The representation $U_t$ contains less information about the future state $S_{t+\Delta}$ than it does about the present state.
    \textbf{(d)} The measurement $U_t$ is a function of $S_t$ and can be learned with a neural network. 
    To extract the information from the current state $S_t$ most predictive of the future state, we optimize the Information Bottleneck objective.
    \textbf{(e)} By distributing bottlenecks, we gain marked interpretability by monitoring the share of optimally predictive information across state variables. 
}
    \label{fig:highlevel}
\end{figure}

\vspace{-1mm}
\section{Approach}
\vspace{-2mm}
Our testbed is a double pendulum (Fig.~\ref{fig:highlevel}a): one of the simplest physical systems to exhibit chaotic behavior~\cite{shinbrot1992chaos}.
We simulated 10,000 trajectories at constant energy (details in the Appendix).

Given a random variable $S_t\sim p(s_t)$ for the system state at time $t$, a measurement $U_t$ is any (possibly stochastic) function of the system state: $U_t \sim p(u_t|s_t)$. 
A measurement process shares mutual information $I(U_t;S_t)$ with the underlying state, defined as the reduction in Shannon's entropy~\cite{shannon1948mathematical} about $S_t$ once $U_t$ is known.
For a continuous variable $S_t$ and without infinite measurement capabilities, the outcome of $U_t$ can only narrow down the possible values of $S_t$, which we visualize in Fig.~\ref{fig:highlevel}b as a region of possible pendulum states. 
We can similarly examine the reduction in entropy about a future state after time $\Delta$ has elapsed, given the same measurement $U_t$. 
Then $I(U_t;S_{t+\Delta})$ serves as an upper bound of the predictive capabilities of any forecasting device given the outcome of the measurement $U_t$.
The passage of time invariably expands our uncertainty about the system state (Fig.~\ref{fig:highlevel}c).

Importantly, different measurements of the state---that is, different measurement processes $U$, not different outcomes $u$ of the same measurement---can have identical $I(U_t;S_t)$ but vary in the information shared with the future state.
We seek the measurement $U_t$ that is most predictive of the future dynamics, for a given allowance of information about the present. 
To find this optimally predictive information, we use the Information Bottleneck~\cite{tishbyIB2000,asoodeh2020bottleneck,creutzig2009pastfutureIB} and optimize over the space of possible measurements.
The following objective maximizes information about the future state and the information about the current state:
\begin{equation} \label{eqn:IB}
    \mathcal{L}_\textnormal{IB} = \beta I(U_t;S_t) - I(U_t;S_{t+\Delta}).
\end{equation}
The parameter $\beta$ controls the bottleneck, restricting the information preserved about the present state.

\begin{figure}
    \centering
    \includegraphics[width=\linewidth]{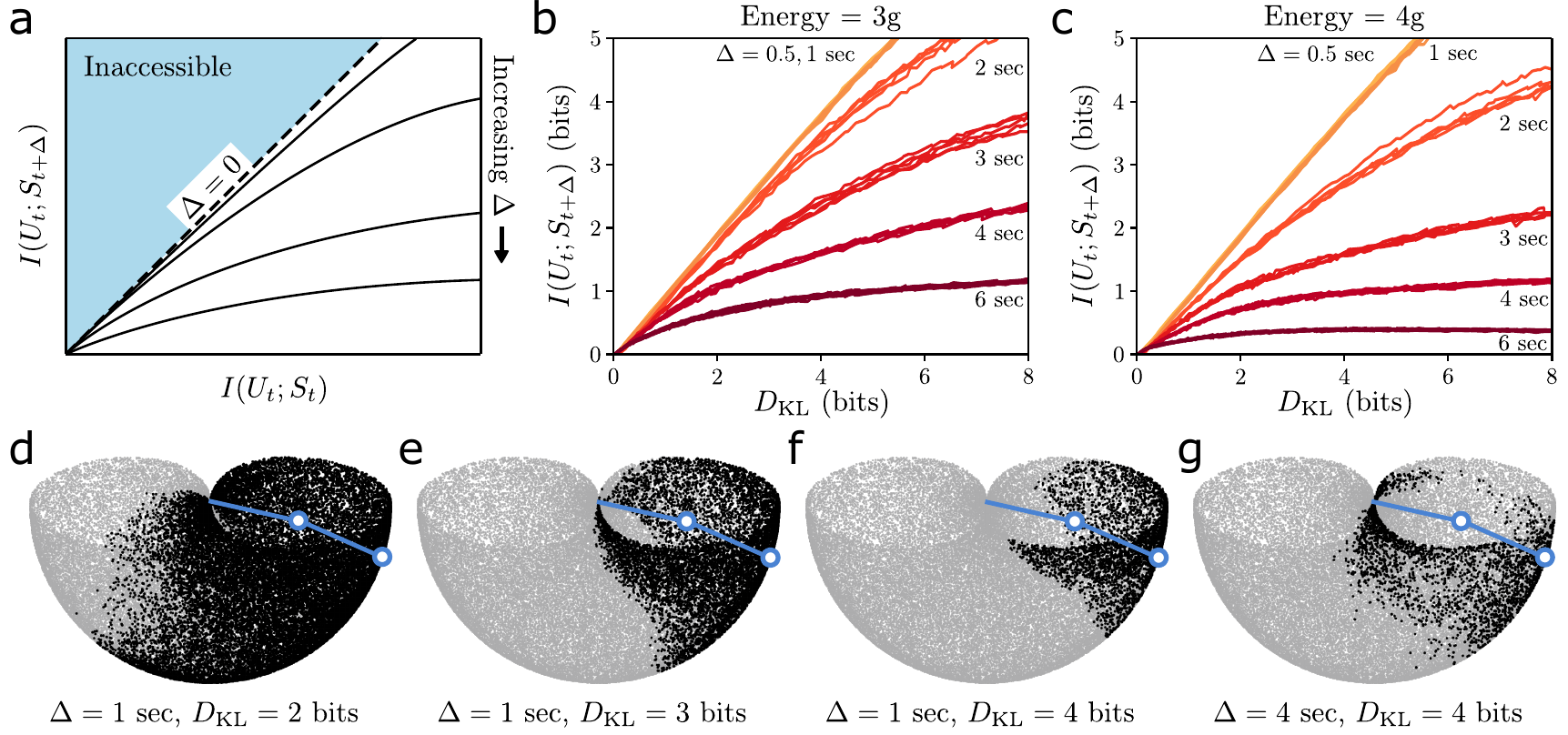}
    \caption{\textbf{Optimally predictive information found by the Information Bottleneck.}
    \textbf{(a)} The information plane compares the amount of information in $U_t$ about the present and future states. Optimally predictive information then traces out trajectories that are determined by the decay of information in the dynamics.
    \textbf{(b, c)} We optimize the variational IB objective and display the trajectories for different time horizons, using the KL-divergence bound discussed in the text, for double pendulums with total energy 3g and 4g, respectively.
    The trajectories from five different splits of the training data are plotted for each $\Delta$.
    \textbf{(d-g)} We visualize learned compressions of a particular pendulum state (overlaid; energy$=$3g) by displaying all other states (dots at the position of the second mass) that have a similar representation $U_t$.
    Gray dots show a random sample over all system states.
}
    \label{fig:infoplane}
\end{figure}

As the bottleneck strength $\beta$ varies, a trajectory in the ``information plane''~\cite{shwartz2017opening} is traced (Fig.~\ref{fig:infoplane}a), which shows the exchange rate of information preserved about the present state versus information predictive of the future state, for different time horizons $\Delta$.
$U_t$ can have no more information about $S_{t+\Delta}$ than it does about $S_t$, so no trajectory can exist above the line with slope 1.
Optimal $U_t$ are as close to this line as possible, and for a chaotic system longer time horizons are further displaced.

Measuring mutual information from data is notoriously tricky~\cite{saxe2019,mcallester2020infolimitations,poole2019variational}.
To be compatible with machine learning, the Variational Information Bottleneck (VIB)~\cite{alemiVIB2016} replaces the mutual information terms of Eqn.~\ref{eqn:IB} with bounds in a framework nearly identical to that of Variational Autoencoders~\cite{vae,betavae}.
The bottleneck of Eqn.~\ref{eqn:IB} is replaced with an upper bound on the mutual information,
\begin{equation}
    I(U_t; S_t) \le D_\textnormal{KL}(p(u_t|s_t)||r(u)).
\end{equation}
\noindent The Kullback-Leibler (KL) divergence---$D_\textnormal{KL}(w(x)||z(x))=\mathbb{E}_{x\sim w(x)}[-\textnormal{log} \ (z(x)/w(x))]$---quantifies the difference between the encoded distribution $p(u_t|s_t)$ and a prior distribution $r(u_t)=\mathcal{N}(0,1)$~\cite{vae}.
As the KL divergence tends to zero, all representations become indistinguishable from the prior and from each other, and therefore uninformative.  
We will use the KL divergence as a proxy for the amount of information contained in the measurement $U_t$.

To estimate the information with the future state, we employ the Noise Contrastive Estimation (InfoNCE) loss used in representation learning as a lower bound for the mutual information in Eqn.~\ref{eqn:IB}, $I(U_t;S_{t+\Delta}) \le \mathcal{L}_\textnormal{InfoNCE}$~\cite{oord2018InfoNCE,poole2019variational} (the loss is reproduced in the Appendix). 
Instead of decoding the representation $U_t$ to a distribution over $S_{t+\Delta}$, which would require discretization of a four-dimensional space, we learn an encoder for the future state and compare $U_t$ to $V_{t+\Delta}$ in a shared representation space.
The degree to which $U_t$ can be matched with its future counterpart in the midst of distractor representations quantifies the amount of information they share.

\vspace{-1mm}
\section{Results and Discussion}
\vspace{-2mm}
As the time horizon increases, we find the expected trend in information plane trajectories (Fig.~\ref{fig:infoplane}b,c): for a given amount of information about the present, less information is recoverable about the future.
What are the optimal measurements $U_t$ for different information allowances and time horizons?
A measurement encodes a particular present state $s$ into a distribution $p(u|s)$ in embedding space; we visualize other states $s^\prime$ that have similar distributions and would therefore be difficult to distinguish given a particular measurement outcome $u$.
We display in Fig.~\ref{fig:infoplane}d-g scatter plots of the positions of the second pendulum mass for all states that are co-embedded at different information rates $D_\textnormal{KL}$.
Specifically, as all states are embedded to multidimensional Gaussian distributions, we use the Bhattacharyya coefficient~\cite{kailath1967Bdistance}---0 for perfectly disjoint and 1 for perfectly matching distributions---to characterize the similarity between distributions, displaying all states with value greater than 0.5.
The measurement process is therefore a clustering of states with similar fates after time $\Delta$.

Visualizing the information content of the optimal measurements leaves much to be desired. 
We necessarily exclude several dimensions of the state, and can do little more than visual inspection.
Instead, we seek a general method to interrogate the optimal information.
We employ the Distributed Information Bottleneck~\cite{dib,aguerri2018DIB,aguerriDVIB2021}, which modifies the optimization objective in ~\ref{eqn:IB} by distributing bottlenecks to each of the state variables (Fig.~\ref{fig:highlevel}e).
While maximizing the predictability of the future state as before, the variables are encoded independently and the sum of information is minimized. 
We acquire a degree of interpretability in the form of information allocation across the measurements, in exchange for less optimal measurements than can be found with the IB.

\begin{figure}
    \centering
    \includegraphics[width=\linewidth]{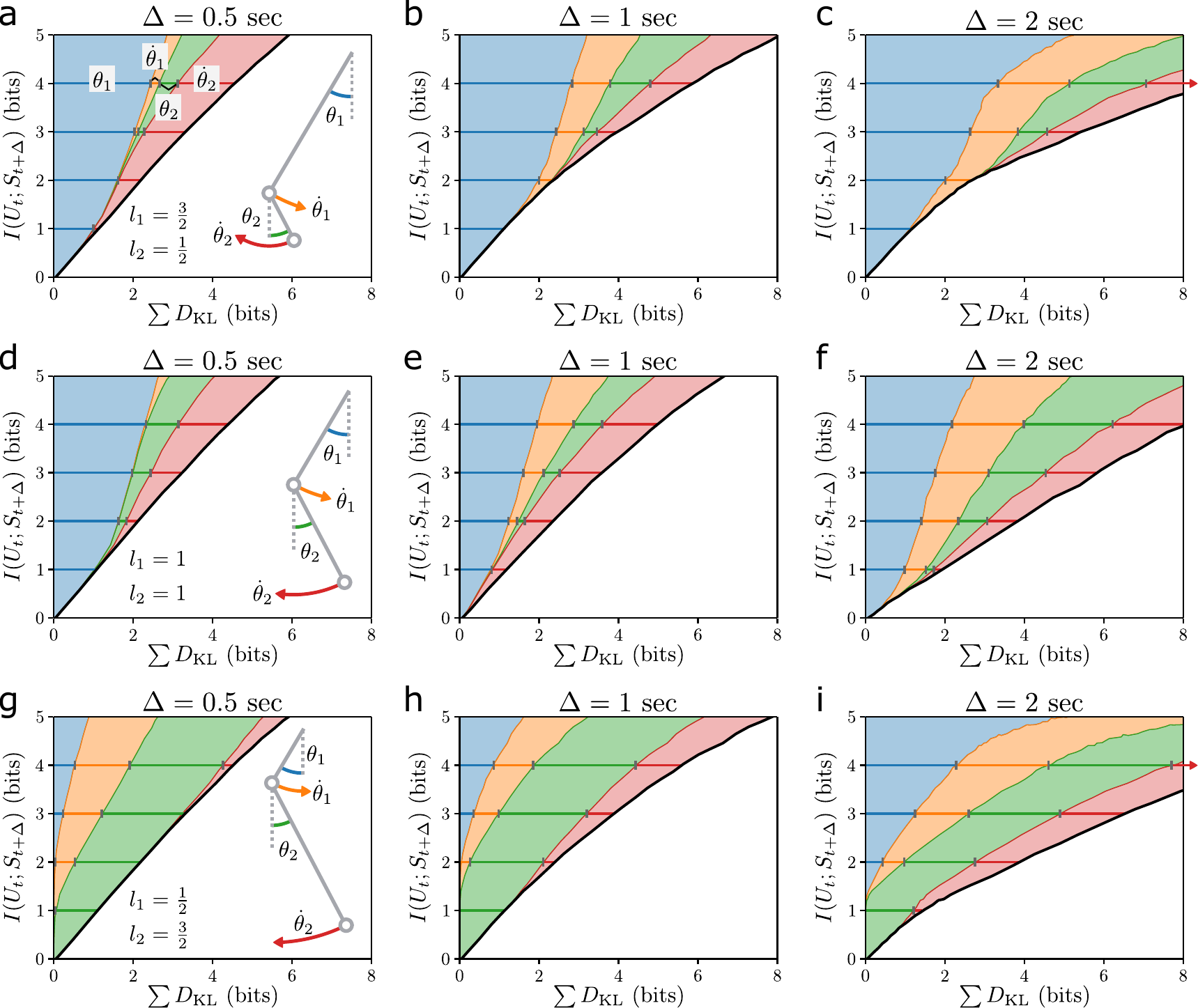}
    \caption{\textbf{The distribution of predictive information over $\theta_1, \dot{\theta}_1, \theta_2,$ and $\dot{\theta}_2$.}
    In each plot, we display the information plane trajectory (black curve) found by optimizing the Distributed Information Bottleneck with a bottleneck on each of the state variables.
    At each point in the trajectory, we display the decomposition of the sum total KL-divergence, an upper bound on the information contained in $U_t$ about the state $S_t$, in terms of the allocation to each of the four state variables.
    The pendulum arm length ratio varies across rows and the time horizon varies across columns.
}
    \label{fig:dib}
\end{figure}

In Fig.~\ref{fig:dib}, the information plane trajectories are decomposed in terms of the KL-divergence contribution from each of the state variables. 
Intuitively, a variable is encoded with a higher KL-divergence if it is more predictive of the future state, thereby granting insight into the relative role each variable plays in the dynamics.
By changing the ratio of the arm lengths in the double pendulum, the variables pertaining to the longer arm become more informative about the future dynamics.

We have developed a framework to analyze the dynamics of a chaotic system through the lens of information theory using machine learning and data.
The Information Bottleneck finds the optimally predictive information in a system state, and the Distributed IB confers interpretability about the role that each variable plays in the dynamics.
The bounds on mutual information facilitate integration with off-the-shelf machine learning models and allow per-sample inspection of information allocation.
The bounds are also the method's primary limitation: careful characterization of the mutual information bounds is necessary for any absolute measurements to be made (e.g., estimates of the Kolmogorov-Sinai or $(\epsilon, \tau)$ entropy).
Nevertheless, even with only relative measurements of information there is a rich source of insight into chaotic dynamics.
\newpage

\section*{Impact statement}
Machine learning struggles with interpretability~\cite{rudin2019stop,rudin2022interpretable}, complicating its application in the natural sciences and many other domains.
In this work we have introduced a machine learning framework whose emphasis is on understanding a dynamical system.
The Distributed Information Bottleneck offers a degree of interpretability in the form of information allocation across components of an input, thereby enhancing the utility of machine learning for the sciences.

{\small
\bibliographystyle{ieee_fullname}
\bibliography{references}

\begin{thebibliography}{10}\itemsep=-1pt

\bibitem{aguerriDVIB2021}
I{\~n}aki~Estella Aguerri and Abdellatif Zaidi.
\newblock Distributed variational representation learning.
\newblock {\em IEEE Transactions on Pattern Analysis and Machine Intelligence},
  43(1):120--138, 2021.

\bibitem{alemiVIB2016}
Alexander~A. Alemi, Ian Fischer, Joshua~V. Dillon, and Kevin Murphy.
\newblock Deep variational information bottleneck.
\newblock {\em arXiv preprint arXiv:1612.00410}, 2016.

\bibitem{asoodeh2020bottleneck}
Shahab Asoodeh and Flavio~P. Calmon.
\newblock Bottleneck problems: An information and estimation-theoretic view.
\newblock {\em Entropy}, 22(11):1325, 2020.

\bibitem{berera2019turbulence}
Arjun Berera and Daniel Clark.
\newblock Information production in homogeneous isotropic turbulence.
\newblock {\em Physical Review E}, 100(4):041101, 2019.

\bibitem{boffetta2002predictability}
Guido Boffetta, Massimo Cencini, Massimo Falcioni, and Angelo Vulpiani.
\newblock Predictability: a way to characterize complexity.
\newblock {\em Physics reports}, 356(6):367--474, 2002.

\bibitem{creutzig2009pastfutureIB}
Felix Creutzig, Amir Globerson, and Naftali Tishby.
\newblock Past-future information bottleneck in dynamical systems.
\newblock {\em Physical Review E}, 79(4):041925, 2009.

\bibitem{aguerri2018DIB}
I{\~n}aki Estella~Aguerri and Abdellatif Zaidi.
\newblock Distributed information bottleneck method for discrete and gaussian
  sources.
\newblock In {\em International Zurich Seminar on Information and Communication
  (IZS 2018). Proceedings}, pages 35--39. ETH Zurich, 2018.

\bibitem{farmer1982informationdimension}
J.~Doyne Farmer.
\newblock Information dimension and the probabilistic structure of chaos.
\newblock {\em Zeitschrift f{\"u}r Naturforschung A}, 37(11):1304--1326, 1982.

\bibitem{gaspard1993noise}
Pierre Gaspard and Xiao-Jing Wang.
\newblock Noise, chaos, and ($\epsilon$, $\tau$)-entropy per unit time.
\newblock {\em Physics Reports}, 235(6):291--343, 1993.

\bibitem{betavae}
Irina Higgins, Loic Matthey, Arka Pal, Christopher Burgess, Xavier Glorot,
  Matthew Botvinick, Shakir Mohamed, and Alexander Lerchner.
\newblock $\beta$-vae: Learning basic visual concepts with a constrained
  variational framework.
\newblock {\em International Conference on Learning Representations (ICLR)},
  2017.

\bibitem{james2014chaosforgets}
Ryan~G. James, Korana Burke, and James~P. Crutchfield.
\newblock Chaos forgets and remembers: Measuring information creation,
  destruction, and storage.
\newblock {\em Physics Letters A}, 378(30-31):2124--2127, 2014.

\bibitem{kailath1967Bdistance}
Thomas Kailath.
\newblock The divergence and {Bhattacharyya} distance measures in signal
  selection.
\newblock {\em IEEE transactions on communication technology}, 15(1):52--60,
  1967.

\bibitem{vae}
Diederik~P. Kingma and Max Welling.
\newblock Auto-encoding variational {B}ayes.
\newblock In {\em International Conference on Learning Representations
  ({ICLR})}, 2014.

\bibitem{mcallester2020infolimitations}
David McAllester and Karl Stratos.
\newblock Formal limitations on the measurement of mutual information.
\newblock In {\em International Conference on Artificial Intelligence and
  Statistics}, pages 875--884. PMLR, 2020.

\bibitem{dib}
Kieran~A. Murphy and Dani~S. Bassett.
\newblock The distributed information bottleneck reveals the explanatory
  structure of complex systems.
\newblock {\em arXiv preprint arXiv:2204.07576}, 2022.

\bibitem{oord2018InfoNCE}
Aaron van~den Oord, Yazhe Li, and Oriol Vinyals.
\newblock Representation learning with contrastive predictive coding.
\newblock {\em arXiv preprint arXiv:1807.03748}, 2018.

\bibitem{poole2019variational}
Ben Poole, Sherjil Ozair, Aaron Van Den~Oord, Alex Alemi, and George Tucker.
\newblock On variational bounds of mutual information.
\newblock In {\em International Conference on Machine Learning}, pages
  5171--5180. PMLR, 2019.

\bibitem{rudin2019stop}
Cynthia Rudin.
\newblock Stop explaining black box machine learning models for high stakes
  decisions and use interpretable models instead.
\newblock {\em Nature Machine Intelligence}, 1(5):206--215, 2019.

\bibitem{rudin2022interpretable}
Cynthia Rudin, Chaofan Chen, Zhi Chen, Haiyang Huang, Lesia Semenova, and Chudi
  Zhong.
\newblock Interpretable machine learning: Fundamental principles and 10 grand
  challenges.
\newblock {\em Statistics Surveys}, 16:1--85, 2022.

\bibitem{saxe2019}
Andrew~M. Saxe, Yamini Bansal, Joel Dapello, Madhu Advani, Artemy Kolchinsky,
  Brendan~D. Tracey, and David~D. Cox.
\newblock On the information bottleneck theory of deep learning.
\newblock {\em Journal of Statistical Mechanics: Theory and Experiment},
  2019(12):124020, dec 2019.

\bibitem{shannon1948mathematical}
Claude~Elwood Shannon.
\newblock A mathematical theory of communication.
\newblock {\em The Bell System Technical Journal}, 27(3):379--423, 1948.

\bibitem{shaw1981}
Robert Shaw.
\newblock Strange attractors, chaotic behavior, and information flow.
\newblock {\em Zeitschrift für Naturforschung A}, 36(1):80--112, 1981.

\bibitem{shinbrot1992chaos}
Troy Shinbrot, Celso Grebogi, Jack Wisdom, and James~A Yorke.
\newblock Chaos in a double pendulum.
\newblock {\em American Journal of Physics}, 60(6):491--499, 1992.

\bibitem{shwartz2017opening}
Ravid Shwartz-Ziv and Naftali Tishby.
\newblock Opening the black box of deep neural networks via information.
\newblock {\em arXiv preprint arXiv:1703.00810}, 2017.

\bibitem{tishbyIB2000}
Naftali Tishby, Fernando~C. Pereira, and William Bialek.
\newblock The information bottleneck method.
\newblock {\em arXiv preprint physics/0004057}, 2000.

\bibitem{wolf1985Lyapunov}
Alan Wolf, Jack~B. Swift, Harry~L. Swinney, and John~A. Vastano.
\newblock Determining {L}yapunov exponents from a time series.
\newblock {\em Physica D: Nonlinear Phenomena}, 16(3):285--317, 1985.

\end{thebibliography}
}

\appendix

\section{Implementation}
\label{app:implementation}
Code and further visualizations are available on the project website, \href{https://distributed-information-bottleneck.github.io}{distributed-information-bottleneck.github.io}.

All experiments were run on a single computer with a 12 GB GeForce RTX 3060 GPU; a single annealing run with 50k steps (without evaluating on the validation set) took about six minutes.

Simulations of the double pendulum were run using \href{https://docs.scipy.org/doc/scipy/reference/generated/scipy.integrate.odeint.html}{\texttt{scipy.integrate.odeint}}, with a timestep $\Delta t =0.001$ sec, masses of $m_1=m_2=1$kg, energy $E \in \{3g, 4g\}$, arm lengths $l_1, l_2 \in \{(1 \textnormal{m}, 1 \textnormal{m}), (1.5\textnormal{m}, 0.5\textnormal{m}), (0.5\textnormal{m}, 1.5\textnormal{m})\}$, and a total simulation time of 100 sec. 
System states were saved every $0.02$ sec.

For each pair of lengths of the pendulum arms $l_1, l_2$, we simulated 10,000 trajectories.
Every starting state had zero kinetic energy and was created by randomly sampling the first pendulum arm's angle uniformly and then solving for the second arm's angle such that the system had the prescribed energy.
The simulation was run for 100 sec, and was discarded if the energy ever deviated by more than 0.001 of the prescribed energy. Data for the first 50 sec was not used in order to minimize the effect of the initialization.

Training hyperparameters and architecture details are shown in Table~\ref{tab:hparams}.
The only tuning that was done involved the annealing: we increased the number of training steps and decreased $\beta_\textnormal{initial}$ until the trajectories were consistent.
The optimization did not appear to be very sensitive to the other parameters (e.g., the architecture specifications).

\begin{table}
\centering
\begin{tabular}{||c c||} 
 \hline
 Parameter & Value \\ 
 \hline\hline
 Nonlinear activation & Leaky ReLU ($\alpha=0.2$) \\
 Encoder MLP architecture (IB) & [128, 128]\\
 Encoders MLP architecture (DIB, one for each of $\theta_1, \dot{\theta}_1, \theta_2, \dot{\theta}_2$) & [128, 128]\\
 Bottleneck embedding space dimension & 32 \\
 Encoder MLP architecture (to shared embedding space) & [256, 256] \\
 Shared embedding space dimension & 64 \\
 Future state encoder MLP architecture & [256, 256] \\
 Positional encoding frequencies & [1, 2, 4, 8, 16, 32, 64, 128] \\
 \hline 
 Batch size & 256 \\
 Optimizer & Adam \\
 Learning rate & $3 \times 10^{-4}$ \\
 $\beta_\textnormal{initial}$ & $5 \times 10^{-4}$ \\
 $\beta_\textnormal{final}$ & $2$ \\ 
 Annealing steps & $5 \times 10^4$\\
 Similarity metric $s(u,v)$ (Eqn.~\ref{eqn:infonce}) & Euclidean squared\\
 \hline
\end{tabular}
\caption{Training parameters for the IB and DIB experiments.}
\label{tab:hparams}
\end{table}

\subsection{InfoNCE loss}
The InfoNCE comparison~\cite{oord2018InfoNCE} is evaluated through the following loss contribution:
\begin{equation}\label{eqn:infonce}
    \mathcal{L}_\textnormal{InfoNCE} = -\sum_i^n \textnormal{log} \frac{\textnormal{exp}(s(u_X^{(i)},u_Y^{(i)})/\tau)}{\sum_j^n \textnormal{exp}(s(u_X^{(i)},u_Y^{(j)})/\tau)},
\end{equation}
\noindent where both sums run over a batch of $n$ examples, $s(u,v)$ is a measure of similarity (e.g., negative Euclidean distance), and $\tau$ acts as an effective temperature.  
Here $X$ and $Y$ refer generally to two random variables being matched; in this work they are $S_t$ and $S_{t+\Delta}$, respectively.

Some intuition about the loss may be gained by viewing it as a classification task.
Eqn.~\ref{eqn:infonce} is equivalent to a cross entropy loss when the correct classification for input $u_X^{(i)}$ is its counterpart $u_Y^{(i)}$, and prediction probabilities are a function of similarities in the shared representation space.
(Specifically, the similarities are used as logits in the prediction, converted to probabilities with a softmax transform.)
In other words, given a pair of instances and a batch of distractor instances, how well do the learned representations allow the pair to be correctly classified?
If the best that can be done is excluding half the possible distractor instances, then there is about one bit of information between $U_X$ and $U_Y$.

\end{document}